\documentclass[10pt,twocolumn,letterpaper]{article}

\usepackage{cvpr}              

\usepackage{graphicx}
\usepackage{amsmath}
\usepackage{amssymb}
\usepackage{booktabs}
\usepackage{float}
\usepackage{makecell}
\usepackage{mathtools}
\usepackage{ragged2e}
\usepackage{pifont}
\newcommand{\cmark}{\ding{51}}%
\newcommand{\xmark}{\ding{55}}%

\usepackage[pagebackref,breaklinks,colorlinks]{hyperref}

\usepackage[capitalize]{cleveref}
\crefname{section}{Sec.}{Secs.}
\Crefname{section}{Section}{Sections}
\Crefname{table}{Table}{Tables}
\crefname{table}{Tab.}{Tabs.}

\def\cvprPaperID{8991} 
\def\confName{CVPR}
\def\confYear{2022}

\begin{document}

\title{Dynamic Sparse R-CNN}

\author{Qinghang Hong\footnotemark[1], Fengming Liu\footnotemark[1], Dong Li, Ji Liu, Lu Tian, Yi Shan \\
 Advanced Micro Devices, Inc., Beijing, China \\
{\tt\small \{d.li, lu.tian, yi.shan\}@amd.com}
}

\maketitle

\renewcommand{\thefootnote}{\fnsymbol{footnote}}
\footnotetext[1]{Equal contribution.}

\begin{abstract}
Sparse R-CNN is a recent strong object detection baseline by set prediction on sparse, learnable proposal boxes and proposal features. In this work, we propose to improve Sparse R-CNN with two dynamic designs. First, Sparse R-CNN adopts a one-to-one label assignment scheme, where the Hungarian algorithm is applied to match only one positive sample for each ground truth. Such one-to-one assignment may not be optimal for the matching between the learned proposal boxes and ground truths. To address this problem, we propose dynamic label assignment (DLA) based on the optimal transport algorithm to assign increasing positive samples in the iterative training stages of Sparse R-CNN. We constrain the matching to be gradually looser in the sequential stages as the later stage produces the refined proposals with improved precision. Second, the learned proposal boxes and features remain fixed for different images in the inference process of Sparse R-CNN. Motivated by dynamic convolution, we propose dynamic proposal generation (DPG) to assemble multiple proposal experts dynamically for providing better initial proposal boxes and features for the consecutive training stages. DPG thereby can derive sample-dependent proposal boxes and features for inference. Experiments demonstrate that our method, named Dynamic Sparse R-CNN, can boost the strong Sparse R-CNN baseline with different backbones for object detection. Particularly, Dynamic Sparse R-CNN reaches the state-of-the-art 47.2\% AP on the COCO 2017 validation set, surpassing Sparse R-CNN by 2.2\% AP with the same ResNet-50 backbone.
\end{abstract}

\section{Introduction}

\begin{figure}
    \centering
    \includegraphics[width=0.5\textwidth]{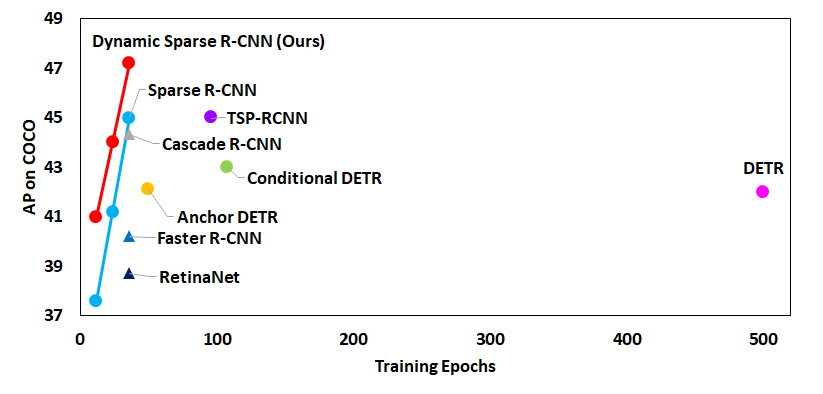}
    \caption{Performance vs. training epochs on the COCO 2017 validation set. All results are reported with single-scale inference using the ResNet-50 backbone. Our Dynamic Sparse R-CNN achieves the state-of-the-art 47.2\% AP with the same 36 training epochs as Sparse R-CNN. Circles: Transformer-based methods. Triangles: CNN-based methods.}
    \label{fig:AP}
\end{figure}

Object detection is a fundamental task in computer vision which aims at predicting a set of objects with locations and corresponding pre-defined categories in a given image.
It has been widely applied in multiple fields including intelligent surveillance and autonomous driving. Object detection has witnessed a rapid development in the recent years, with varying feature extraction backbones from convolutional neural network (CNN) \cite{simonyan2014very,freund2003efficient,Szegedy_2016_CVPR,he2016resnet} to Transformer \cite{dosovitskiy2020vit,liu2021Swin} and varying detection pipeline designs \cite{Ren2015Faster,liu2016ssd,redmon2018yolov3,tian2019fcos,carion2020end, sun2021sparse}. The detectors can mainly be divided into one-stage, two-stage and multi-stage methods according to the regression times. One-stage detectors \cite{tian2019fcos,liu2016ssd} directly predict the regression targets and categories of objects in a given image without the refinement step. Two-stage detectors \cite{Ren2015Faster,girshick2015fast,dai2016rfcn,li2017lightheadrcnn} first generate a limited number of candidate proposals for foreground (e.g., region proposal network (RPN)) and then pass the proposals to the detection network to refine the location and category. Multi-stage detectors \cite{cai18cascadercnn} would refine the location and category multiple times with improved performance but often require large computation overhead. One-stage methods generally can be divided into anchor-based and anchor-free detectors. Anchor-based detectors \cite{liu2016ssd,redmon2018yolov3,lin2017focal} design dense pre-defined anchors, tile anchors across the image, and then directly predict category and refine the coordinates of anchors. However, manual anchor configurations could be sub-optimal for the final performance. Anchor-free detectors \cite{tian2019fcos,kong2020foveabox} are proposed to overcome this issue by removing the anchor design. 
They typically use center points or regions inside ground truth to define positive proposals and predict offsets to obtain final bounding boxes. 

Recently, Transformer-based detectors\cite{carion2020end,sun2021sparse,wang2021anchor,meng2021-CondDETR} have been proposed by formulating object detection as a set prediction problem using the Transformer encoder and decoder architecture. These methods replace anchor mechanisms with a small number of learnable object queries, which can model the relationships between objects and global image context to output the final predictions. Hungarian algorithm is used to find a bipartite matching between ground truths and predictions based on the combined loss of classification and regression. The label assignment in these detectors is a one-to-one way where only one single detection matches one ground truth during training. 

Motivated by existing CNN-based methods using many-to-one label assignment schemes \cite{lin2017focal,tian2019fcos,ge2021ota}, We assume that assigning multiple positives to a GT can optimize the proposals more efficiently and can promote the detector training for better performance. Thus, we propose dynamic label assignment (DLA) with many-to-one matching based on the optimal transport algorithm for the strong baseline of Sparse R-CNN. We also adopt gradually increasing positive samples assigned to GTs in the iterative stages of Sparse R-CNN. Since each stage produces refined proposal boxes and features for the next one, we expect to constrain the matching between GTs and prediction boxes to be stricter in the early stages and looser in the later stages owing to the increasing precision of predictions in the sequential stages. Moreover, in Sparse R-CNN, the object queries (i.e., proposal boxes and proposal features) are learnable during training but remain fixed for different images during inference. Motivated by dynamic convolution \cite{chen2020dynamic}, we propose dynamic proposal generation (DPG) to provide better initial proposal boxes and features in the first iterative stage. Compared to fixed proposals, DPG can aggregate multiple parallel proposal experts which are sample-dependent and output dynamic proposals for inference. We name our method as Dynamic Sparse R-CNN, which reaches the state-of-the-art 47.2\% AP on the COCO 2017 validation set, surpassing the Sparse R-CNN baseline by a large margin of 2.2\% AP with the same ResNet-50 backbone (Figure \ref{fig:AP}).

Our main contributions can be summarized as follows. (1) We point out that many-to-one label assignment in Transformer-based detection is more reasonable and effective than the one-to-one scheme. We apply the optimal transport assignment method into Sparse R-CNN and assign gradually increasing positive samples to GTs in the iterative stages. (2) We design a dynamic proposal generation mechanism to learn multiple proposal experts and assemble them for generating dynamic proposal boxes and features for inference. (3) We integrate the two dynamic designs into Sparse R-CNN and the resulting Dynamic Sparse R-CNN detector obtains a large AP gain of 2.2\%, reaching the state-of-the-art 47.2\% AP on the COCO validation set with ResNet-50.

\section{Related Work}
 
\subsection{General Object Detection}
CNN-based detectors have achieved great progress owing to the development of various feature extraction backbones and pipeline designs. One-stage detectors directly predict the location and associated categories of object in a given image without region proposal and refinement components, including anchor-based\cite{liu2016ssd,lin2017focal,redmon2018yolov3} and anchor-free\cite{tian2019fcos,kong2020foveabox} methods. Two-stage detectors \cite{Ren2015Faster,dai2016rfcn,li2017lightheadrcnn} first generate a fixed number of proposal for foreground with region proposal network (RPN) and then pass the proposals to the detection network for refining the locations and categories of objects. 

Recently, Transformer-based detectors\cite{carion2020end,zhu2021deformable,wang2021anchor,meng2021-CondDETR} utilizes Transformer encoder and decoder architecture to re-formulate the object detection as a set prediction problem. They design a small number of learnable object queries to model the relations between objects and the global image context, and have shown impressive performance. Object queries in decoders are a required component of DETR \cite{carion2020end} (7.8\% AP drops without them). Conditional DETR \cite{meng2021-CondDETR} proposes a conditional spatial query for fast training convergence. Anchor DETR \cite{wang2021anchor} proposes a query design based on anchor points and achieve near performance to DETR with less training time. Sparse R-CNN \cite{sun2021sparse} proposes learnable proposal boxes and proposal features, and pass the RoI features extracted on the feature map (based on proposal boxes) and associated proposal features to the iterative structure (i.e., dynamic head) for prediction.

\subsection{Label Assignment}
Label Assignment plays a prominent part in modern object detectors. Anchor-based detectors\cite{liu2016ssd,lin2017focal,Ren2015Faster} usually adopt IoU at a certain threshold as the assigning criterion. For example, RetinaNet defines the anchors having IoU score higher than 0.5 as positive samples and others as negative samples. YOLO detectors\cite{redmon2017yolo9000,redmon2018yolov3} only adopt the anchor having the max IoU score associated to the ground-truth as the positive sample and such label assignment is a one-to-one matching method. Anchor-Free detectors\cite{tian2019fcos,kong2020foveabox,zhou2019objects} define center points or shrinking center regions of ground truth as positives and take others as negatives. ATSS\cite{zhang2019bridging} indicates that the essential difference between anchor-based and anchor-free detectors is label assignment. It proposes an adaptive training sample selection method which divides positive and negative samples according to statistical characteristics of object. PAA\cite{kim2020probabilistic} proposes a probabilistic anchor assignment method by modeling the distribution of joint loss for positive and negative samples as the Gaussian distribution. OTA\cite{ge2021ota} formulates the label assignment as an optiomal transport problem by defining ground truths and background as supplier and defining anchors as demander, and then employs Sinkhorn-Knopp Iteration to efficiently optimize the problem. Transformer-based detectors\cite{carion2020end,zhu2021deformable,sun2021sparse,wang2021anchor,meng2021-CondDETR} formulate object detection as a set prediction problem and treat label assignment between ground truths and object queries as a bipartite matching. Hungarian algorithm is used to optimize the one-to-one matching between ground truths and object queries by minimizing the global loss. In this paper, we assume that one-to-one label assignment is sub-optimal in Transformer-based detectors and explore a dynamic label assignment with many-to-one matching for Sparse R-CNN inspired by OTA \cite{ge2021ota}. 

\subsection{Dynamic Convolution}
Dynamic convolution\cite{chen2020dynamic} is a technique that dynamically combines multiple convolution kernels with learnable sample-dependent weights to enhance the representation capability of the model. Temperature annealing in softmax can help improve both the training efficiency and final performance.
CondConv \cite{NEURIPS2019_CondConv} proposes conditionally parameterized convolutions, which learn specialized convolution kernels for each input image. It combines multiple convolution kernels with weights generated with sub-net using sigmoid transformation to construct a image-specified convolution kernel. DyNet \cite{2020DyNet} designs several dynamic convolution neural networks based on dynamic convolution including Dy-mobile, Dy-shuffle and Dy-ResNet, etc. In this work, we analyze that the fixed proposal boxes and features in Sparse R-CNN for different inputs during inference is sub-optimal and inflexible. Motivated by dynamic convolution, we improve Sparse R-CNN by generating dynamic sample-dependent proposals during inference.

\section{Proposed Approach}

\subsection{Revisit Sparse R-CNN}
\begin{figure*}[t!]
\begin{center}
\begin{tabular}{@{}cc@{}}
\includegraphics[width = 0.4\linewidth]{{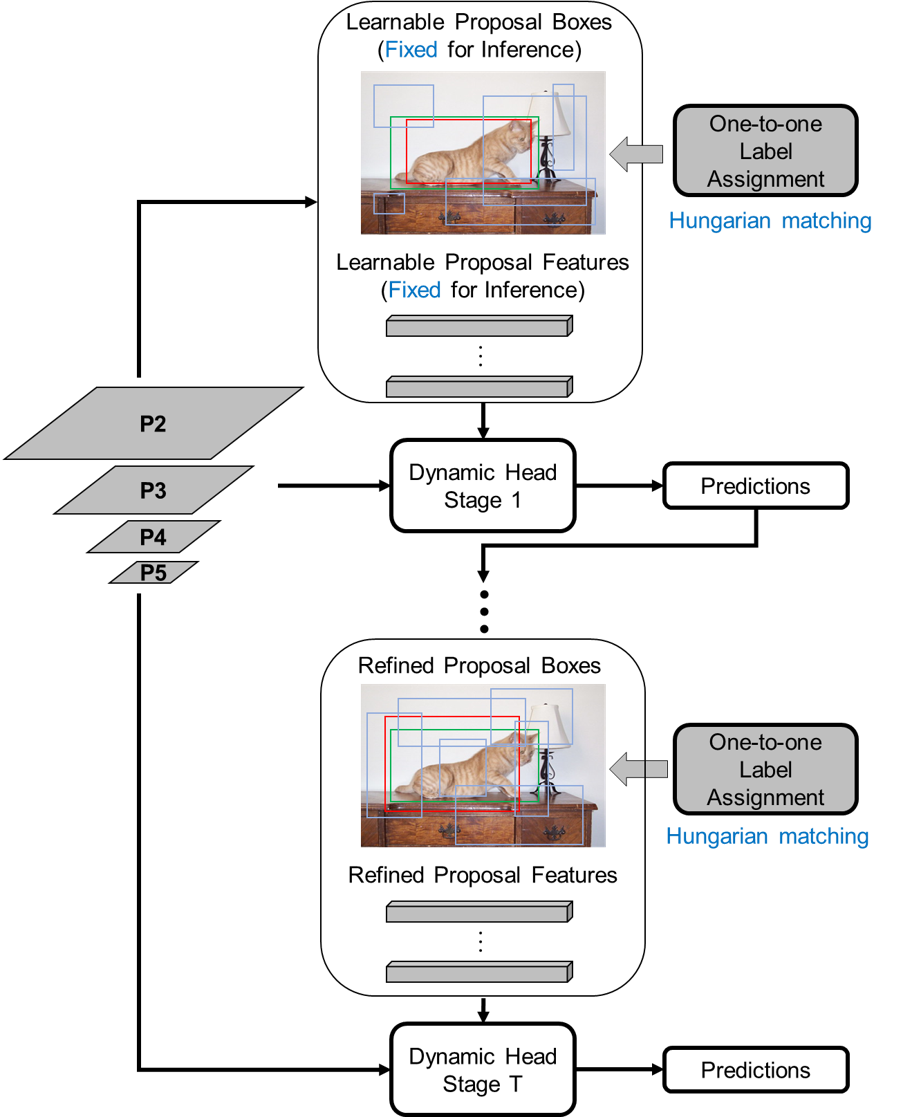}} \hspace{5mm} & 
\includegraphics[width = 0.4\linewidth]{{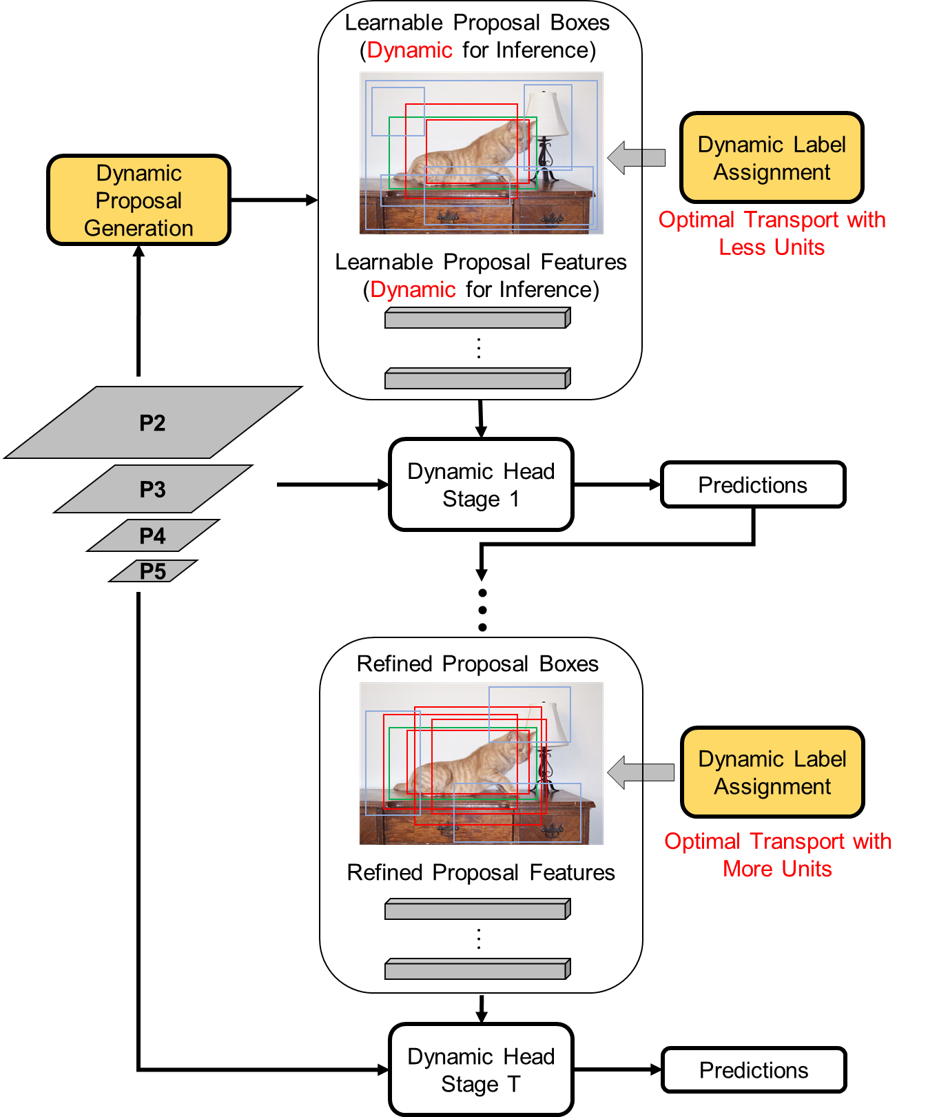}} \vspace{3mm}\\
{(a) Sparse R-CNN} \hspace{5mm} & 
{(b) Dynamic Sparse R-CNN}
\end{tabular}
\end{center}
\vspace{-3mm}
\caption{Comparisons with (a) the Sparse R-CNN baseline and (b) our Dynamic Sparse R-CNN. Sparse R-CNN uses one-to-one label assignment optimized by Hungarian algorithm and fixed proposal boxes / features during inference. Dynamic Sparse R-CNN improve Sparse R-CNN with two dynamic designs. First, we adopt dynamic many-to-one label assignment optimized by the optimal transport algorithm with unit increasing strategy. Second, we propose dynamic proposal generation to generate sample-dependent proposal boxes and features.}
\label{figure:overview}
\end{figure*}

Sparse R-CNN \cite{sun2021sparse} is a recent strong object detection baseline by set prediction on a sparse set of learnable object proposals. It uses an iterative structure (i.e., dynamic head) to gradually produce and refine the predictions. The input of each iterative stage consists of three parts: FPN features extracted by the backbone, proposal boxes and proposal features. The output includes the predicted boxes, the corresponding classes and object features of the boxes. The predicted boxes and object features output by one stage are respectively used as the refined proposal boxes and proposal features to the next stage. Proposal boxes are a small fixed set of region proposals ($N_p \times 4$), indicating the potential locations of the objects. Proposal features are latent vectors ($N_p \times C$) to encode the instance characteristics (e.g., pose and shape). In Sparse R-CNN, proposal boxes are learned during training and fixed for inference. Sparse R-CNN applies the set-based loss to produce a bipartite matching between predictions and ground
truth objects, which uses one-to-one matching with the Hungarian algorithm. Figure \ref{figure:overview} (a) illustrates the design of Sparse R-CNN.

We analyze two main limitations of Sparse R-CNN as follows. First, Sparse R-CNN adopts one-to-one matching between the detection predictions and the ground truths, which is likely to sub-optimal and inefficient for training. Second, the learned proposal boxes and proposal features in Sparse R-CNN represent the statistics of the training set, which are not adaptive for a specific test image. In our work, we devise two modifications to improve Sparse R-CNN. Figure \ref{figure:overview} gives the overview of our method and we introduce algorithm details in the following sections.


\subsection{Dynamic Label Assignment}
In Sparse R-CNN, the Hungarian algorithm is used for one-to-one matching, where each ground truth is matched to one predicted box. We assume that such one-to-one matching is likely to be sub-optimal. Assigning multiple positives to a GT can optimize the proposals more efficiently and promote the detector training. 

To implement many-to-one matching, we follow the CNN-based method \cite{ge2021ota} and apply the optimal transport assignment (OTA) in Transformer. Specifically, OTA is a formulation that explores how the detection boxes should be matched to ground truths. The formulation treats the ground truths as suppliers to provide quota for assignment, and treats detection boxes as demanders to seek for assignments. The background class is also formulated as a supplier that provides default assignment.

Mathematically, suppose that we have $m$ ground truths in an image and each provides $s_i = k$ assignments, which are referred as \textit{units}. Each of $n$ detection boxes tries to get an unit and a successful matching is referred as a positive assignment. The background provides $s_i = n-k*m$ units to fulfill detection boxes that are not assigned to any ground truth, which is referred as negative assignments. The optimization target can be defined as follows.
\begin{equation}
\footnotesize
\begin{split}
    &\mathop{\min}_{\pi} \sum_{i=1}^{m} \sum_{j=1}^{n} C(i,j) * \pi(i,j),\\  
    &\text{s.t.} \sum_{i=1}^{m} \pi(i,j) = 1, \text{ } \sum_{j=1}^{n} \pi(i,j) = s_i, \text{ } \sum_{i=1}^{m} s_i = n, \\
    &\pi(i,j) > 0, i = 1,2, ..., m, j = 1,2, ..., n, \\
    &C(i,j) = \begin{cases}
       L_{cls}(i, j) + \alpha * L_{reg}(i,j), \text{positive assignment} \\
       L_{cls}(background, j), \text{negative assignment}
       \end{cases}
\end{split}
\label{eq: ota}
\end{equation}
where $i$ is the index of ground truth, $j$ is the index of detection boxes ($j=1,...n$), $\alpha$ is a coefficient balancing the classification and regression losses. The cost of each positive assignment is the sum of the classification loss $L_{cls}$ and regression loss $L_{reg}$, while the cost of each negative assignment is only the classification loss. $\pi(i,j)$ represents the matching result to be optimized between ground truth $i$ and detection box $j$. 

The number of units $k$ offered by each supplier can be fixed or dynamic. Following the Dynamic $k$ Estimation method in \cite{ge2021ota}, our work dynamically estimates the $k$ value based on the IoU between the predictions and the ground-truth boxes. In this strategy, top $q$ IoU values for each ground truth are selected and summed up (and converted to an integer) as the estimation for the $k$ value. 
Based on the optimal transport theory for label assignment ($\sum_{i=1}^{m} \pi(i,j) = 1$ in Eq. \ref{eq: ota}), each proposal (i.e., demander) only needs one unit of label provided by GT (i.e., supplier). Thus, one proposal will not be assigned to different GTs. The Dynamic $k$ Estimation method generally holds $k<q$. Suppose that $m$ is the number of GTs and $N_p$ is the number of total proposals, if $m \times k > 80\% \times N_p$, we will reduce $k$ by a same scaling factor for each GT to ensure at least 20\% negative assignments. 

\textbf{Unit Increasing Strategy.}
Sparse R-CNN adopts an iterative architecture to gradually increase the precision of predictions. We present a simple unit increasing strategy to promote the training of iterative structure. When the predictions of dynamic head are not correct enough in the early stage, we expect the suppliers (GT) to provide a small number of units, which constrain the matching to be stricter. When the predictions of dynamic head become more correct in the later stage, we gradually relax the constraints to let the suppliers (GT) provide a larger number of units for matching. The simple unit increasing strategy can be defined as follows.
\begin{gather}
    k^* = k - 0.5 * (T - t), t=1,2, ...T
    \label{equ:strategy 2}
\end{gather}
where we use the default number of iteration stages ($T=6$) in our method.





\begin{figure}[t!]
\begin{center}
\begin{tabular}{@{}c@{}}
\includegraphics[width = 0.9\linewidth]{{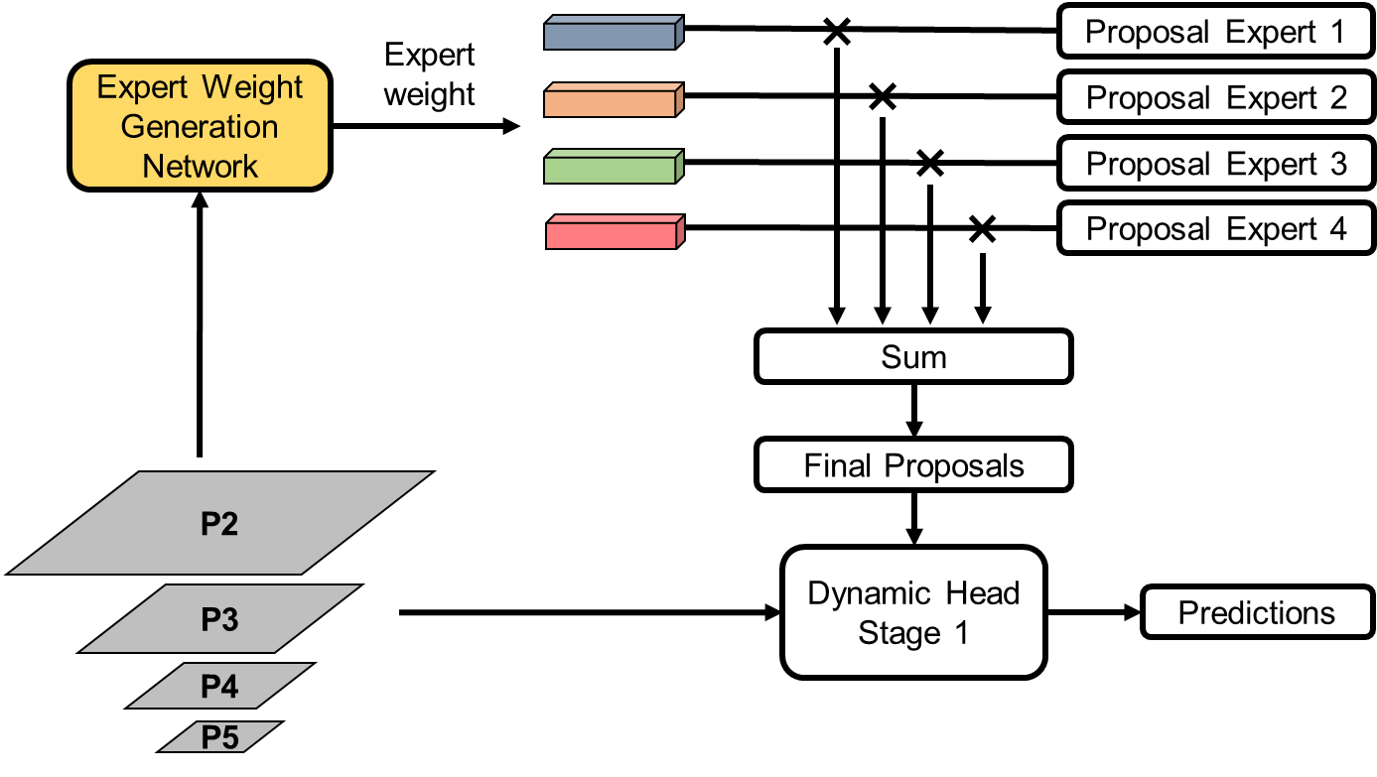}} \\
{(a) Dynamic Proposal Generation} 
\vspace{5mm} \\
\includegraphics[width = 0.9\linewidth]{{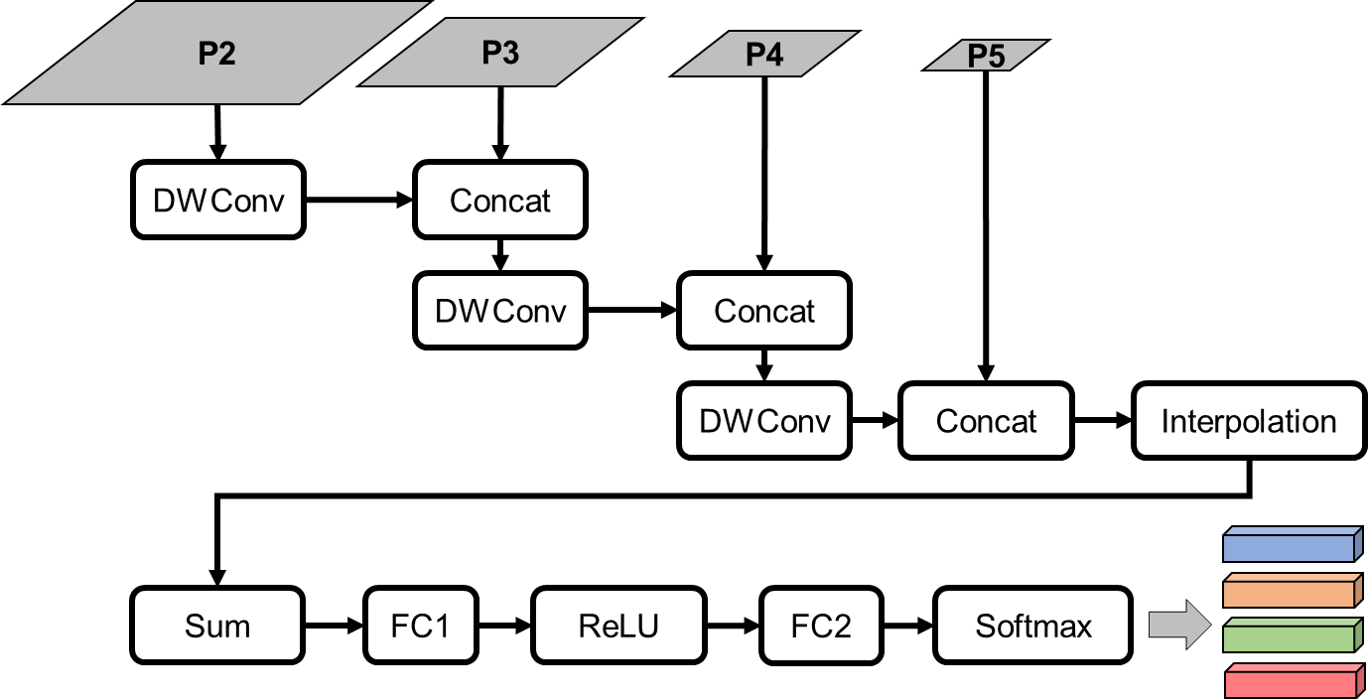}} \\
{(b) Staircase Structure} 
\end{tabular}
\end{center}
\vspace{-5mm}
\caption{Illustrations of the proposed (a) Dynamic Proposal Generation (DPG) module and (b) staircase structure in DPG to produce expert weights.}
\label{figure:dpg}
\end{figure}

\subsection{Dynamic Proposal Generation}

In Sparse R-CNN, a set of $N_p$ proposal boxes and $N_p$ proposal features are fed into the dynamic head together with the features extracted from the FPN backbone ($P_2$ to $P_5$). These proposals are learnable during training but fixed for different images during inference. Motivated by dynamic convolution, we propose to generate dynamic proposal boxes and features with respect to the input image to improve performance. In our design (Figure \ref{figure:dpg} (a)), proposal boxes / features are a linear combination of $N_e$ distinct sets of proposal boxes / features, and each set is referred to as an \textit{expert}. The coefficients ( referred to as expert weights) to combine the experts are generated by an expert weight generation network (Figure \ref{figure:dpg} (b)). Our DPG module can be formulated as follows.
\begin{equation}
\begin{array}{c}
\mathcal{P}^b_{o} = \sum_{i=1}^{N_e} \mathcal{P}^b_i * W_i \\
\mathcal{P}^f_{o} = \sum_{i=1}^{N_e} \mathcal{P}^f_i * W_i \\
(W_1, W_2, ..., W_{N_e}) = G(\mathcal{F}) \\
\end{array}
\end{equation}
where $\mathcal{P}^b_i$ indicates the output dynamic proposal boxes, $\mathcal{P}^f_i$ indicates the output dynamic proposal features, $W_i$ is the proposal expert weight learned by the expert weight generation network $G$, $\mathcal{F}$ indicates the features extracted from the FPN backbone ($P_2$ to $P_5$).

\textbf{Staircase Structure.} Our expert weight generation network follows the basic design of dynamic convolution structure, as shown in Figure \ref{figure:dpg} (b). We also use the temperature annealing operation (tao) in softmax to control the expert weights and make the training process more effective. We build a staircase architecture to aggregate the features from different pyramid levels. The $P_2$ to $P_5$ features descend in scale: the width and height of $P_i$ is $1/2$ of that of $P_{i-1}$. Depth-wise convolution with $3\times 3$ kernel and stride of 2 is applied to the concatenation of $P_i$ and the output by the previous level, which keeps the number of channels and downscales the intermediate features. Finally, the concatenated data is interpolated into a $4C \times 30 \times30$ feature map ($C=256$ for each pyramid level). Then, the $4C$ channels are fused by summation and the resulting $30\times30$ feature map is flattened to two FC layers. The size of the first FC is $900\times1500$ and the second is $1500\times(N_eN_p)$. We build $N_e=4$ experts and use $N_p=300$ proposal boxes / features in our method.

All experts as well as the expert weight generation network are trained. During inference, the weight generation network takes the FPN features as input and generates the weight for each expert. Then the final proposal boxes and features are obtained by linear combinations of experts. 



\section{Experiments}

\textbf{Datasets.} All experiments are conducted on the COCO 2017 dataset \cite{lin2014microsoft}. The training split contains about 118k samples and the validation split contains about 5k samples. This dataset labels 80 different categories of objects, which are collected from natural scenarios. We use the standard MS COCO AP as the main evaluation criterion. 

\textbf{Training Details.} The basic training setting follows Sparse R-CNN. We use the pre-trained networks (e.g., ResNet-50 \cite{he2016resnet}) on ImageNet \cite{deng2009imagenet} with 5 FPN levels as backbone. 
During training, we use AdamW optimizer \cite{loshchilov2017decoupled} and the weight decay is set to 0.0001. We train the model with batch size of 16 for 36 epochs. The initial learning rate is $2.5 \times 10^{-5}$ and scaled with 0.1 at 27-th and 33-th epoch. Xavier initialization \cite{glorot2010understanding} is applied to newly added layers. We follow Sparse R-CNN to adopt the same multi-scale training procedure by resizing the input images such that the shortest side is at least 480 and at most 800 pixels while the longest at most 1333. 
Following Sparse R-CNN, we adopt the iterative structure with 6 stages for training. Our experiments are conducted on 4 Nvidia A100 GPUs and training of Dynamic Sparse R-CNN takes around 37 hours with the ResNet-50 backbone.

\textbf{Inference Details.} 
For inference, 300 boxes and the associated scores are output as predictions. The score of each box is the probability that the box contains an object. No post-processing on these boxes is needed during inference. In our dynamic label assignment based on OTA, non-maximum suppression (NMS) is applied with threshold of 0.7.

\begin{table*}[t!] \small
  \begin{center}
  \begin{tabular}{l | c | c | c|c|c|c|c|c}
    \toprule
     Methods &   Backbone  & Train Epochs  &AP & AP$_{50}$ & AP$_{75}$ & AP$_S$ & AP$_M$ & AP$_L$ \\
    \midrule
    \textit{CNN-based Detectors:} & & & & & & & \\
    Faster R-CNN \cite{wu2019detectron2} & ResNet-50 & 36 & 40.2 &61.0 &43.8 &24.2 &43.5 &52.0 \\
    
    Faster R-CNN \cite{wu2019detectron2} & ResNet-101 & 36 & 42.0 &62.5& 45.9 &25.2& 45.6& 54.6 \\
    
     RetinaNet  \cite{wu2019detectron2} & ResNet-50 & 36 & 38.7 &58.0& 41.5& 23.3& 42.3 &50.3 \\
    
     RetinaNet  \cite{wu2019detectron2} & ResNet-101 & 36 & 40.4 & 60.2& 43.2 & 24.0 &44.3& 52.2 \\
     Cascade R-CNN  \cite{wu2019detectron2} & ResNet-50 & 36 & 44.3 & 62.2 & 48.0 & 26.6& 47.7& 57.7 \\
    
     ATSS\cite{zhang2019bridging} & ResNet-101 & 24 & 43.5 & - &- & - & - & - \\
     PAA\cite{kim2020probabilistic} & ResNet-101 & 24 & 44.6 & - &- & - & - & - \\
     OTA\cite{ge2021ota} & ResNet-50 & 12 & 40.7 & 58.4 &44.3&23.2&45.0&53.6 \\
     
    
    \midrule
    \textit{Transformer-based Detectors:} & & & & & & & \\
    DETR \cite{carion2020end} & ResNet-50 & 500 & 42.0 & 62.4 & 44.2 & 20.5 & 45.8 & 61.1   \\
    DETR \cite{carion2020end} & ResNet-101 & 500 & 43.5 & 63.8 & 46.4 & 21.9 &48.0 & 61.8   \\
    DETR \cite{carion2020end} & ResNet-101-DC5 & 500 & 44.9& 64.7 &47.7& 23.7& 49.5& 62.3   \\
    Conditional DETR\cite{meng2021-CondDETR}  & ResNet-50&108 &43.0 & 64.0 & 45.7 & 22.7 & 46.7 & 61.5 \\
    Conditional DETR\cite{meng2021-CondDETR} & ResNet-101 & 108& 44.5 & 65.6 & 47.5 & 23.6 & 48.4 & 63.6 \\
    Conditional DETR\cite{meng2021-CondDETR} & ResNet-101-DC5 & 108& 45.9 & 66.8 & 49.5 & 27.2 & 50.3 &63.3 \\
    Anchor DETR \cite{wang2021anchor} & ResNet-50 & 50&42.1 &63.1 & 44.9 & 22.3 & 46.2 & 60.0   \\
    Anchor DETR\cite{wang2021anchor} & ResNet-101 & 50 &43.5 & 64.3 & 46.6 & 23.2 & 47.7 & 61.4 \\
        Anchor DETR\cite{wang2021anchor} & ResNet-101-DC5 & 50 &45.1& 65.7 &48.8 &25.8 &49.4 &61.6 \\
    Sparse\_R-CNN \cite{sun2021sparse} & ResNet-50 &36 &45.0 & 63.4 & 48.2 & 26.9& 47.2 & 59.5  \\  
   Sparse\_R-CNN \cite{sun2021sparse} & ResNet-101 & 36&46.4 &64.6& 49.5& 28.3& 48.3& 61.6  \\  
    TSP-RCNN \cite{sun2021rethinking} &ResNet-50 & 96& 45.0 & 64.5 & 49.6 & 29.7 & 47.7& 58.0  \\
    TSP-RCNN \cite{sun2021rethinking} &ResNet-101 & 96& 46.5 & 66.0& 51.2 &29.9& 49.7& 59.2 \\
    \midrule
    \textit{Ours:} & & & & & & & \\
    Dynamic Sparse R-CNN & ResNet-50 & 36 & 47.2 & 66.5 & 51.2 & 30.1 & 50.4 & 61.7 \\
    Dynamic Sparse R-CNN & ResNet-101 & 36 & 47.8 & 67.0 & 52.0 & 31.0 & 51.1 & 62.2 \\

    \bottomrule
  \end{tabular}
  \end{center}
    \caption{Detection performance comparisons (\%) on the COCO 2017 validation set. 
    }
    \label{table:sota}
\end{table*}

\begin{table*}
    \begin{center}
    \begin{tabular}{l|c|c|c|c|c|c}
    \toprule
    Setting & AP & $AP_{50}$ & $AP_{75}$ & $AP_s$ & $AP_m$ & $AP_l$  \\
    \midrule
    Baseline & 45.0 & 63.4 & 48.2 & 26.9 & 47.2 & 59.5 \\
    + DPG, w/o staircase & 45.3 & 63.2 & 49.5 & 28.8 & 48.2 & 59.1 \\
    + DPG, w/ staircase & 45.7 & 63.9 & 50.0 & 28.8 & 48.2 & 59.8 \\
    + DPG, + DLA, dynamic $q$=8, w/o unit increasing strategy & 46.0 & 65.0 & 49.9 & 28.7 & 49.2 & 61.1 \\
    + DPG, + DLA, dynamic $q$=8, w/ unit increasing strategy & 47.2 & 66.5 & 51.3 & 30.1 & 50.4 & 61.7 \\
    \bottomrule
    \end{tabular}
    \end{center}
    \caption{Effect of each algorithmic component of our method.}
    \label{tab:Dynamic Proposal + OTA Result}
\end{table*}

\subsection{Comparisons to the State-of-the-Arts}
\textbf{Comparisons to Transformer-based Detectors.} Table \ref{table:sota} compares our Dynamic Sparse R-CNN with the state-of-the-art Transformer-based object detection methods which are mostly related to our method. The results show that Dynamic Sparse R-CNN outperforms not only the original Sparse R-CNN, but also the other improved DETR methods, such as Conditional DETR and Anchor DETR. For example, with the same ResNet-50 backbone, our work surpasses Conditional DETR by 4.2\% AP and Anchor DETR by 5.1\% AP. Equipped with a larger ResNet-101 backbone, we also obtain improved performance compared to prior methods by a large margin. On the other hand, we only train the network for 36 epochs (same as the Sparse R-CNN baseline), which are significantly shorter than other Transformer-based detectors. We also evaluate our method on the COCO test-dev set. Our Dynamic Sparse R-CNN achieves 47.2\% AP with ResNet-50 and 47.9\% with ResNet-101, which surpasses TSP-RCNN with ResNet-101 (46.6\%).

\begin{figure}
    \centering
    \includegraphics[width=0.45\textwidth]{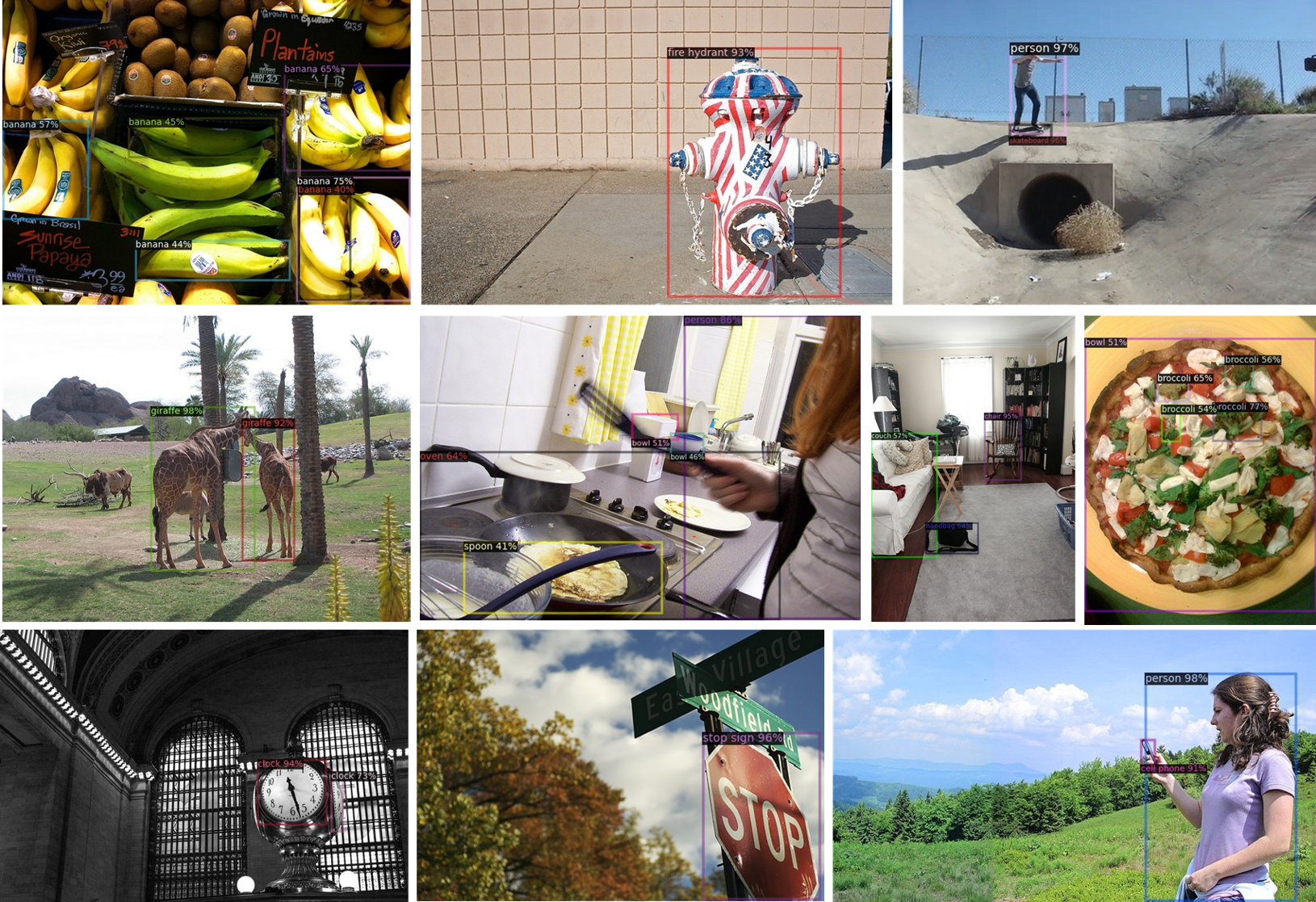}
    \caption{Visualization of sampled detection results by Dynamic Sparse R-CNN with the ResNet-50 backbone.}
    \label{fig:Result visualization}
\end{figure}

\begin{table*} \small
    \begin{center}
    \begin{tabular}{c|c|c|c|c|c|c|c|c|c|c}
    \toprule
    Backbone & Matcher & unit & loss & {\makecell[c]{unit increasing \\ strategy}} & AP & $AP_{50}$ & $AP_{75}$ & $AP_s$ & $AP_m$ & $AP_l$  \\
    \midrule
    R50 & Hungarian & \xmark & \xmark & \xmark & 45.0 & 63.4 & 48.2 & 26.9 & 47.2 & 59.5 \\
    R50 & OTA & fixed $k$=1 & \xmark & \xmark & 44.7 & 64.9 & 48.0 & 28.2 & 46.9 & 59.3 \\
    R50 & OTA & fixed $k$=2 & \xmark & \xmark & 45.9 & 65.1 & 49.8 & 28.8 & 48.6 & 60.9 \\
    R50 & OTA & fixed $k$=3 & \xmark & \xmark & 45.9 & 65.2 & 50.0 & 28.6 & 48.6 & 61.0 \\
    R50 & OTA & dynamic $q$=8 & \xmark & \xmark & 46.1 & 64.6 & 50.1 & 27.9 & 49.2 & 61.9 \\
    R50 & OTA & dynamic $q$=8 & two losses & \xmark & 46.1 & 65.2 & 50.0 & 29.4 & 49.7 & 60.9 \\
    R50 & OTA & dynamic $q$=8 & two losses & \cmark & 46.7 & 65.9 & 50.9 & 29.8 & 49.8 & 61.3  \\
    \bottomrule
    \end{tabular}
    \end{center}
    \caption{Effect of different matchers. Dynamic proposal generation is not used in this ablation experiment.}
    \label{tab:OTA Result}
\end{table*}

\textbf{Comparisons to CNN-based Detectors.} We also compare our Dynamic Sparse R-CNN with the state-of-the-art CNN-based methods. With the same 3$\times$ training scheduler (i.e., 36 epochs), our method outperforms Faster R-CNN, RetinaNet and Cascade R-CNN. The methods of ATSS, PAA and OTA also explore improved many-to-one label assignment schemes which are related to our DLA. Our Dynamic Sparse R-CNN obtain superior performance compared to these methods with the same backbone, e.g., surpassing OTA by 6.5\% AP with ResNet-50 and PAA by 3.2\% AP with ResNet-101 on the COCO validation set.

\textbf{Qualitative Results.} Figure  \ref{fig:Result visualization} visualizes sampled detection results by our Dynamic Sparse R-CNN. Our method can detect objects correctly with varying scales, appearances, etc.

\subsection{Ablation Study}

\textbf{Contributions from Algorithmic Components.}
We conduct ablation experiments to examine the contributions from each algorithmic components. As shown in Table \ref{tab:Dynamic Proposal + OTA Result}, the dynamic proposal generation design boosts AP by 0.7 points with the staircase structure to aggregate features from multiple pyramid levels. In particular, both the $AP_{75}$ and $AP_{s}$ values witness an enhance for nearly 2 points, demonstrating that that DPG helps the model to perform better in a more strict IoU criterion and detecting small objects. The intuition behind this improvement is that the DPG helps to provide a more diverse range of proposal boxes and features to the dynamic head for better predictions. Our staircase structure can better utilize the FPN features for generating expert weights. Without staircase structure, the FPN features are directly interpolated into the $30\times30$ feature maps and concatenated to be fed into the first FC layer. The results show that this staircase structure brings 0.4\% AP gain. By applying many-to-one label assignment based on OTA, we can boost the performance from 45.7\% to 46.0\%. In this setting, the units are set based on the same Dynamic $k$ Estimation method for all the iteration stages. We find that our simple unit increasing strategy can further improve the performance, reaching 47.2\% AP with a single model. These results demonstrate the effectivenss of our designs of DLA and DPG.

\begin{table}[t!]
    \begin{center}
    \begin{tabular}{c|c|c|c|c|c|c}
    \toprule
    $q$ & AP & $AP_{50}$ & $AP_{75}$ & $AP_s$ & $AP_m$ & $AP_l$  \\
    \midrule
    4 & 46.7 & 66.0 & 51.1 & 31.5 & 50.1 & 60.5 \\
    5 & 46.7 & 66.2 & 50.9 & 30.6 & 49.8 & 61.1 \\
    6 & 46.7 & 66.0 & 50.9 & 30.2 & 50.0 & 60.7 \\
    7 & 46.4 & 65.7 & 50.4 & 30.2 & 49.5 & 60.7 \\
    8 & 47.2 & 66.5 & 51.3 & 30.1 & 50.4 & 61.7 \\
    9 & 46.1 & 65.2 & 50.1 & 29.0 & 49.5 & 60.6 \\
    \bottomrule
    \end{tabular}
    \end{center}
    \caption{Effect of $q$ in Dynamic $k$ Estimation with unit increasing strategy and dynamic proposal generation.}
    \label{tab:q in dynamic k estimation with dynamic proposal}
\end{table}

\textbf{Effect of Different Matchers.}
As shown in Table \ref{tab:OTA Result}, OTA matchers with fixed $k$ values ($k=2,3$) gives a 0.9-point lift of AP compared to the baseline. The OTA matcher with $q=8$ in Dynamic $k$ Estimation brings a higher increase of 1.1 points, which demonstrates the effectiveness of using dynamic $k$. The unit increasing strategy further enhances AP to 46.7\%, indicating that this simple design is effective. In addition, the OTA matcher with $q=8$ and the unit increasing strategy brings a nearly 3-point increase in terms of both $AP_{75}$ and $AP_{s}$. The intuition behind the significant increase is that our dynamic many-to-one matching scheme produces more diverse options of prediction boxes to match a ground truth. This scheme especially favors the detection of small objects.

\textbf{Effect of $q$.} 
As shown in Table \ref{tab:q in dynamic k estimation with dynamic proposal}, we try different choices of $q$ in Dynamic $k$ Estimation and find $q=8$ works best. It is noted that all the results in Table \ref{tab:q in dynamic k estimation with dynamic proposal} outperforms the one-to-one matching baseline (45.0\%), which validate the effectiveness of our dynamic many-to-one matching scheme.

\textbf{Effect of Number of Experts.} As shown in Table \ref{tab:Number of experts Ablation}
, we try different numbers of experts and use 4 experts as default in our method.

\begin{table}[t!]
    \begin{center}
    \begin{tabular}{c|c|c|c|c|c|c}
    \toprule
    {\makecell[c]{\#Experts}} & AP & $AP_{50}$ & $AP_{75}$ & $AP_s$ & $AP_m$ & $AP_l$  \\
    \midrule
    3 & 45.4 & 63.4 & 50.0 & 28.6 & 48.4 & 59.6\\
    4 & 45.7 & 63.9 & 50.0 & 28.8 & 48.2 & 59.8\\
    5 & 45.3 & 63.2 & 49.5 & 27.6 & 47.8 & 60.0\\
    \bottomrule
    \end{tabular}
    \end{center}
    \caption{Effect of the number of experts. Dynamic label assignment is not used in this ablation experiment.}
    \label{tab:Number of experts Ablation}
\end{table}

\section{More Analysis}
\begin{figure}
    \centering
    \includegraphics[width=0.5\textwidth]{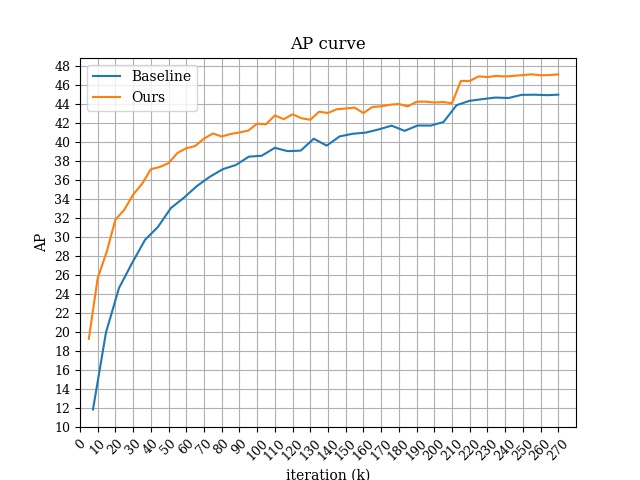}
    \caption{Comparisons of AP curves between Sparse R-CNN and Dynamic Sparse R-CNN.
}
    \label{fig:AP curve}
\end{figure}

Figure \ref{fig:AP curve} compares the detailed training curve of the AP values between Sparse R-CNN and Dynamic Sparse R-CNN. We observe that our Dynamic Sparse R-CNN outperforms the baseline throughout the training iterations. The results further validate the non-trivial design of DLA and DPG. 

\begin{figure}
    \centering
    \includegraphics[width=0.5\textwidth]{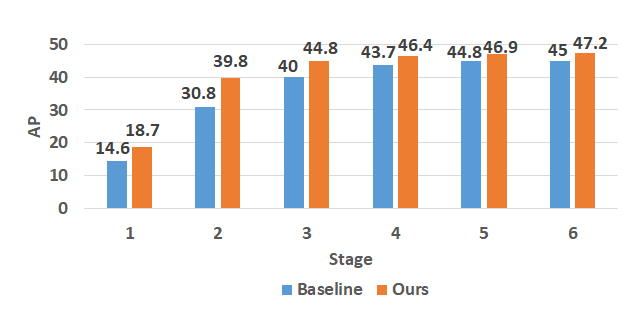}
    \caption{Comparisons of per-stage results between Sparse R-CNN and Dynamic Sparse R-CNN.}
    \label{fig:Per-stage result}
\end{figure}

Figure \ref{fig:Per-stage result} compares the per-stage AP values between Sparse R-CNN and Dynamic Sparse R-CNN. The AP value of each stage is improved by at least 2 points using our method. This indicates that DLA and DPG actually contribute to the training of each iteration stage. We note that DPG is imposed for the first stage only, it helps produce better initial proposal boxes and features and could benefit the consecutive stages.
Moreover, we find that Dynamic Sparse R-CNN already can achieve 46.4\% AP using 4 stages, outperforming the baseline (45.0\%) using 6 stages. The results show that our method can accelerate the convergence in the iterative structure.



\section{Limitations}

The parameter size and computation cost of our detector are slightly larger than the  Sparse R-CNN baseline. Sparse R-CNN has 77.8M parameters and costs 23.28 GFLOPs, while our Dynamic Sparse R-CNN has 81.0M parameters and costs 23.30 GFLOPs. It indicates that our expert weight generation network just introduces marginal memory and computation overhead. Our Dynamic Sparse R-CNN takes 37 hours on 4 A100 GPUs for training, whereas Sparse R-CNN takes 29 hours on the same devices. The training time could be optimized further.

\section{Conclusion}
In this work, we propose Dynamic Sparse R-CNN by introducing two dynamic designs to improve Sparse R-CNN. We point out that one-to-one label assignment method is sub-optimal for matching between object queries and ground truths in Transformer-based detectors. Based on optimal transport algorithm, we implement many-to-one label assignment and design a simple but effective unit increasing strategy for performance boost. We also propose a dynamic proposal generation mechanism to aggregate multiple learned experts to derive better initial proposal boxes and features. Such mechanism is motivated by dynamic convolution and produces dynamic input-dependent proposals for better detection performance. Our Dynamic Sparse R-CNN is well-motivated and reaches the state-of-the-art 47.2\% AP with ResNet-50 on COCO. We expect our method can inspire new insights for object detection and consider applying our idea to more Transformer-based detectors as future work.

{\small
\bibliographystyle{ieee_fullname}
\bibliography{egbib}
}

\end{document}